\documentclass{article}

\PassOptionsToPackage{numbers, compress}{natbib}

     \usepackage[preprint]{neurips_2019}

\usepackage[utf8]{inputenc} 
\usepackage[T1]{fontenc}    
\usepackage{hyperref}       
\usepackage{url}            
\usepackage{booktabs}       
\usepackage{amsfonts}       
\usepackage{nicefrac}       
\usepackage{microtype}      
\usepackage{cite}
\usepackage{epsfig}
\usepackage{graphicx}
\usepackage{subfigure}
\usepackage{multirow}
\usepackage{color}
\usepackage{array}
\usepackage{colortbl} 
\usepackage{arydshln}
\usepackage{authblk}
\title{Masked Non-Autoregressive Image Captioning}

\author{%
	Junlong Gao$^{1}$~~~~Xi Meng$^{2}$~~~~Shiqi Wang$^{5}$~~~~Xia Li$^{1}$~~~~Shanshe Wang$^{3,4}$\thanks{Corresponding author}~~~~Siwei Ma$^{3,4}$~~~~Wen Gao$^{3,4}$\\
	$^{1}$Peking University Shenzhen Graduate School\\
	$^{2}$National Key Lab for Novel Software Technology, Nanjing University\\
	$^{3}$Institute of Digital Media, Peking University~~~~~$^{4}$Peng Cheng Laboratory\\
	$^{5}$Department of Computer Science, City University of Hong Kong, Hong Kong
}

\begin{document}
	
	\maketitle
	
	\begin{abstract}
		Existing captioning models often adopt the encoder-decoder architecture, where the decoder uses autoregressive decoding to generate captions, such that each token is generated sequentially given the preceding generated tokens. However, autoregressive decoding results in issues such as sequential error accumulation, slow generation, improper semantics and lack of diversity. Non-autoregressive decoding has been proposed to tackle slow generation for neural machine translation but suffers from multimodality problem due to the indirect modeling of the target distribution. In this paper, we propose masked non-autoregressive decoding to tackle the issues of both autoregressive decoding and non-autoregressive decoding. In masked non-autoregressive decoding, we mask several kinds of ratios of the input sequences during training, and generate captions parallelly in several stages from a totally masked sequence to a totally non-masked sequence in a compositional manner during inference. Experimentally our proposed model can preserve semantic content more effectively and can generate more diverse captions.
	\end{abstract}
	
	\section{Introduction}
	Image captioning aims at generating natural captions automatically for images, where most recent works adopt the encoder-decoder framework \citep{Vinyals2015Show,xu2015show,anderson2017bottom,lu2017knowing}. In general, an encoder encodes images into visual features, and a decoder decodes the image features to generate captions. Most models use autoregressive decoders, such as LSTM \citep{Hochreiter1997Long}, which generate each token conditioned on the sequence of generated tokens in a sequential manner. This process is not parallelizable, which results in slow generation, and is prone to sequential error accumulation once the preceding tokens are inappropriate during inference \citep{bengio2015scheduled,lamb2016professor}. This process is unable to model the inherent hierarchical structures of natural languages \citep{Manning1999Foundations}, which makes autoregressive decoders heavily favor the frequent n-grams in the training data \citep{Bo2017Towards}. Hence, autoregressive decoders also risk improper semantics and lack of diversity. While non-autoregressive decoding \citep{Gu2018Non} has been proposed for neural machine translation (NMT) to tackle slow generation of autoregressive decoding, but another problem, dubbed as "multimodality problem", due to the indirect modeling of the true target distribution, is inevitably introduced. 
	
	In this paper, we propose masked non-autoregressive decoding for image captioning to address the problems of autoregressive decoding and non-autoregressive decoding. In the masked non-autoregressive decoding, motivated by the masking strategy in BERT \citep{Devlin2018BERT}, during training we randomly mask the input sequences with certain ratios to train a masked language model to address multimodality problem, and during inference the model generates captions parallelly in several stages from a totally masked sequence to a totally non-masked sequence in a compositional manner, which is faster and can generate more semantically correct captions than autoregressive decoding. It is a first visual then linguistic generation process, i.e. the salient visual information is generated in early stages and linguistic information assists to form the final caption with the masked language model. The experimental results suggest that masked non-autoregressive decoding can generate captions with richer semantics and more diversity compared to autoregressive decoding.
	\section{Background}
	
	\subsection{Autoregressive decoding}
	Given an image $I$, image captioning aims to generate a token sequence
	$A = \left\{ {{a_1},{a_2},...,{a_T}} \right\}$, ${a_t} \in \mathbb{A}$, where $\mathbb{A}$ is the dictionary and $T$ is denoted as the sequence length. In image captioning, the standard encoder-decoder architecture encodes an image $I$ to the visual feature ${I_F}$ using an encoder, and decodes ${I_F}$ to a token sequence $A$ using a decoder.
	In autoregressive decoding, a decoder generates a token conditioned on the sequence of tokens generated thus far. The decoder predicts a sequence starting with token [BOS] ${a_0}$ and ending with token [EOS] ${a_T}$. It corresponds to the word-by-word nature of human language generation and effectively captures the conditional distribution of real captions. Formally, given an image $I$, the model factors the distribution over possible output sentences $A$ into a chain of conditional probabilities with a left-to-right structure:
	\begin{equation}
	{\pi _\theta }\left( {A\left| {{I_F}} \right.} \right){\rm{ = }}\prod\limits_{t = 1}^T {{\pi _\theta }\left( {{a_t}\left| {{a_{0:t - 1}},{I_F}} \right.} \right)}
	\label{eq1}
	\end{equation}
	where ${\pi _\theta }\left( {{a_t}\left| {{a_{0:t - 1}},{I_F}} \right.} \right)$ is a probability distribution of the token ${a_t}$ given the generated tokens $\left\{ a_{0:t - 1} \right\}$ and the visual feature ${I_F}$.
	During training, given a ground-truth sequence $A^*=\left\{ {a_1^*,a_2^*,...,a_T^*} \right\}$, the model parameters $\theta $ are trained to minimize the conditional cross entropy loss as follows: 
	\begin{equation}
	L\left( \theta  \right) =  - \sum\limits_{t = 1}^T {\log {\pi _\theta }\left( {{A^ * }\left| {{I_F}} \right.} \right)}
	\label{eq2}
	\end{equation}
	The mainstream autoregressive decoders include recurrent neural networks (RNNs) \citep{Vinyals2015Show}, masked convolutional neural networks (CNNs) \citep{Aneja2017Convolutional} and transformer~\citep{Vaswani2017Attention}. RNNs adopt hidden states to model the temporal state transformation of input tokens, and thus models cannot be trained parallelly. Since the entire target sequence is known during training, the preceding target tokens can be used in the calculation of later conditional probabilities and their corresponding losses. CNNs and transformer can take advantage of this parallelism during training.
	However, during inference, at each step a token is generated and then fed into the decoder to predict the next token. Therefore, all autoregressive decoders must remain sequential rather than parallel during inference. Moreover, sequential decoding is prone to copying tokens from the training data to enhance the grammatical accuracy, which easily causes semantic error and is lack of diversity regarding the generated captions in image captioning.
	
	\subsection{Non-autoregressive decoding}
	In order to tackle slow generation of autoregressive decoders during inference, non-autoregressive decoder is proposed to produce the whole sentence parallelly in only one step, allowing an order of magnitude lower latency. Non-autoregressive decoding was first proposed by  \citep{Gu2018Non} to deal with NMT. The model has an explicit likelihood function as follows and is trained using independent cross entropy loss
	\begin{equation}
	{\pi _\theta }\left( {A\left| {{I_F}} \right.} \right){\rm{ = }}{\pi _\theta }\left( {T\left| {{I_F}} \right.} \right)\prod\limits_{t = 1}^T {{\pi _\theta }\left( {{a_t}\left| {{I_F}} \right.} \right)} 
	\label{eq3}
	\end{equation}
	However, this naive approach is unable to achieve desirable results, since complete conditional independence results in the poor approximation to the true target distribution, which is dubbed as "multimodality problem". To tackle this issue, they utilized "fertility" and several complicated tricks to indirectly model the true target distribution, which is still inferior to autoregressive decoding. 
	\section{Masked non-autoregressive image captioning}
	
	In this paper, in order to tackle the problems of autoregressive decoding and non-autoregressive decoding, we introduce a novel image captioning model Masked Non-autoregressive Image Captioning (MNIC), which generates captions through masked non-autoregressive decoding. MNIC can produce the entire output captions parallelly with richer semantics in constant stages, the number of which is much smaller than the sequence length $T$.
	
	\subsection{Model architecture}
	The model architecture of MNIC is modified from the traditional transformer \citep{Vaswani2017Attention}, where it composes of an encoder and decoder. In this work, the encoder and decoder are devised as follows. 
	
	In image captioning, the input of an encoder is an image, which is a low-level semantic data rather than a high-level semantic sequence in NMT. Therefore, we directly adopt a different encoder, that firstly extracts feature maps from CNNs or object features from object detection models, then maps the dimension of input visual features to the model dimension of the decoder using a Multi-Layer Perceptron (MLP).
	As for the decoder, we adopt the original decoder of transformer with necessary modifications. In order to generate captions non-autoregressively and parallelly, we remove the autoregressive mask in the masked multi-head attention layer of each decoder layer, such that self-attention layers in the decoder allow each position to attend to all positions in the decoder. 
	
	\subsection{Masked non-autoregressive decoding}
	\begin{figure}
		\centering
		\includegraphics[scale=0.8]{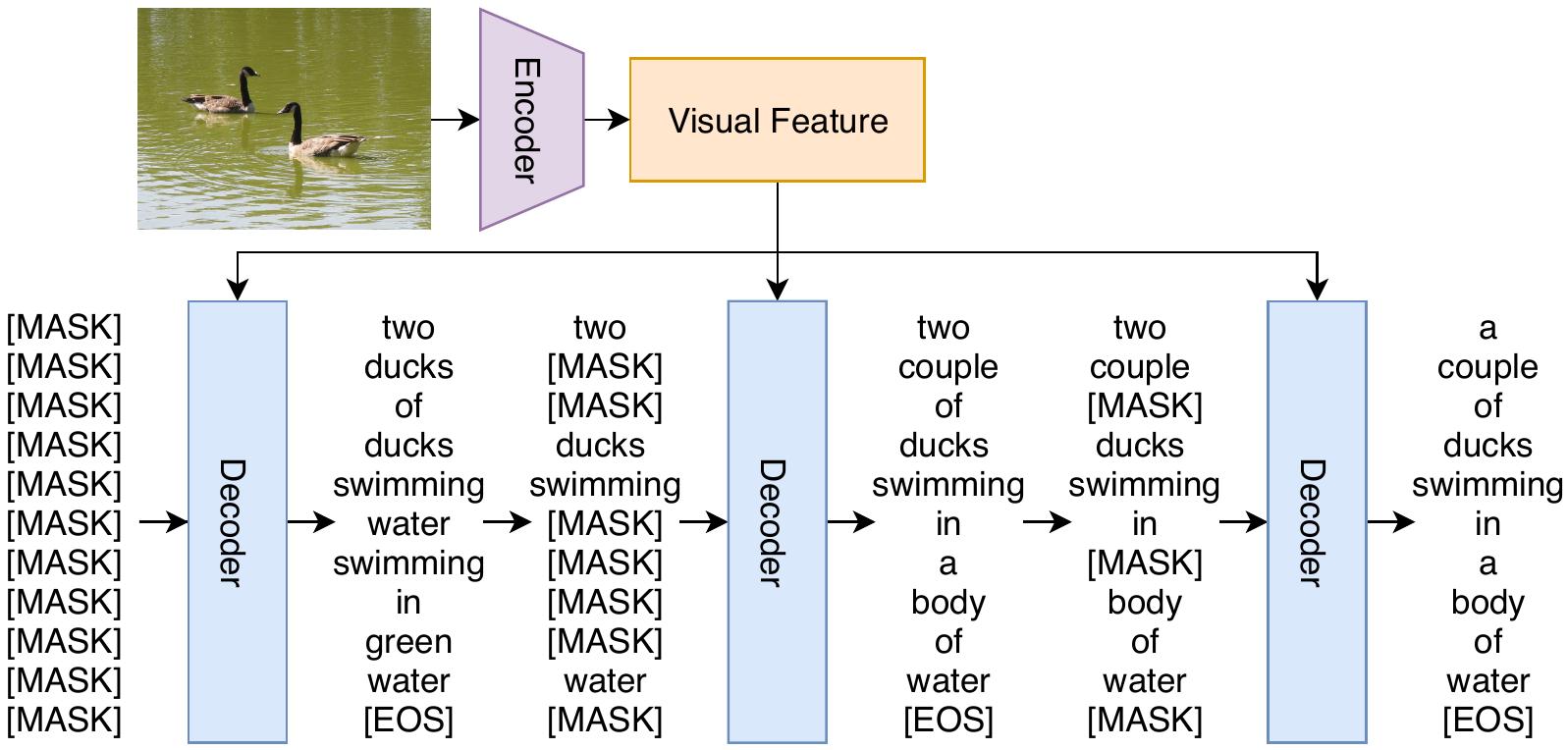}
		\caption{An overview of MNIC paradigm under the setting of $K=3$ and $R={\rm{\{ }}0.2, 0.6, 1.0 {\rm{\} }}$ masking ratio set. The final caption is generated through $K$ stages from an entirely masked sequence and visual feature generated by the encoder. In each stage, a masked sequence is fed into the decoder with visual feature, and then the output sequence is selectively masked again and fed to the decoder in the next stage until the last one. It is worth noting that the masking ratios of these input masked captions in $K$ stages are $1.0$, $0.6$ and $0.2$ respectively, which correspond to the masking ratio set $R$ during training. The decoders in blue boxes are identical.}
		\label{fig1}
	\end{figure}

	In masked non-autoregressive decoding, during the training process we randomly mask the input sequence with $K$ kinds of ratios to train a masked bidirectional language model, and during the inference process the model generates captions parallelly in $K$ stages from a totally masked sequence to a totally non-masked sequence in a compositional manner, which is faster and can generate more semantically correct captions than autoregressive decoding. Herein, the masked tokens are replaced with [MASK] token, the $K$ kinds of ratios form a ratio set $R$, and $a_{1:T}^r$ is denoted as the masked input sequence, which is generated by randomly masking the target sequence $a_{1:T}$ with the masking ratio $r$. Therefore, given the visual features \(I_F\) and $a_{1:T}^r$, the masked language model has a likelihood function of the entire tokens $a_{1:T}$ as follows which can be trained using cross entropy loss, 
	\begin{equation}
	{\pi _\theta }\left( {A\left| {{I_F}} \right.} \right){\rm{ = }}{\pi _\theta }\left( {T\left| {{I_F}} \right.} \right)\prod\limits_{t = 1}^T {{\pi _\theta }\left( {{a_t}\left| {{a_{1:T}^r}, {I_F}} \right.} \right)} 
	\label{eq4}
	\end{equation}
	
	When $r$ is $1.0$, the model is forced to train a fully non-autoregressive decoder, and thus the non-autoregressive decoding is a special case of our model. When $r$ is small, the model is a bidirectional language model, which is proved in BERT \citep{Devlin2018BERT} to be more powerful than the unidirectional model. 
	
	Here, we use an example to further clarify the training and inference phases. Assume that $K=3$ and the set of $K$ kinds of ratios can be $R={{\rm{\{ }}0.2, 0.6, 1.0 {\rm{\} }}}$. During training, the input sequence containing these $K$ ratios [MASK] tokens respectively will be fed into the model, and the masked positions are randomly selected. During inference, as depicted in Fig.~\ref{fig1}, firstly the entirely masked sequence (ratio $1.0$) will be fed into the model to generate a coarse caption, then the coarse caption is processed by replacing $0.6$ less informative tokens with [MASK] token to generate a masked coarse caption, and the masked coarse caption is sent into the model to generate a finer caption. Finally the finer caption is masked again with $0.2$ tokens to generate a masked finer caption, and the masked finer caption is sent into the model to generate a complete sequence. The distributions of the input data of training and inference phases are consistent. Therefore, an entire token sequence can be generated parallelly in only $K$ stages, which is similar to the traditional non-autoregressive decoders but at a cost of more $K-1$ times of computation. However, using masked input sequence can train a bidirectional language model to directly model the true target distribution and thus address multimodality problem of non-autoregressive decoding, in analogous to BERT.
	
	However, there are several differences between MNIC and BERT \citep{Devlin2018BERT}. BERT has only one kind masking ratio and is relatively small ($0.15$) to train deep bidirectional representation for natural language understanding. However, MNIC is designed for generating complete token sequences from scratch for natural language generation. As such, various masking ratios uniformly ranging from $\left[ {{\rm{0.0, 1.0}} } \right]$ have been adopted.
	
	We also choose some ratios of random words to replace [MASK] token or ground-truth token. In our experiment, since the length of caption is relatively short, we simply replace a word in each non totally masked input sequence with a random word. Using random words can enhance the contextual representation of tokens, and improve the robustness of the inference process through introducing noise tokens during training, as the model easily generates wrong tokens in early stages of inference.
	
	\subsection{Analysis}
	Here we provide more analyses and discussions regarding what the model learns with different ratios for masking sequences and what the model predicts in different stages during inference. We further discuss the intrinsic differences between autoregressive and masked non-autoregressive decoding.
	
	During training, when the input data are masked with the larger ratio it is difficult to force the model to predict complete and well organized captions given few clues. One potential reason lies in that expressions are protean in training data, especially that the captions are full of twists and turns. The model will learn some words that can easily be inferred from visual features and some words of high frequency in early stages. Therefore, it is difficult to learn a non-autoregressive decoder for tasks that have no input data with well organization in high-level semantic (e.g., image captioning). By contrast, it could be easier for tasks that offer the input data with well organization in high-level semantic and meanwhile aim to generate sequences to just restate the input data (e.g., machine translation). With the input data masked with the smaller ratio, the model will be given more clues to reconstruct the entire sequences and trained as a bidirectional language model, such that it is believed to generate a grammatically correct sentence.
	
	During inference, the masked non-autoregressive decoding process well reflects what the model learns during training. In early stages the model tends to generate a caption containing tokens of high frequency (e.g., "a", "on") and salient visual clues (e.g., objects, color nouns and important verbs) in images with poor language organization, and in later stages the model can generate a semantically and grammatically correct captions through adopting the trained bidirectional language model to select the most suitable words to concatenate both sides of sub-sequence. As illustrated in Fig.~\ref{fig1}, the output sequence of the first stage is disordered in terms of grammatical rules, but includes some important keywords such as "two", "ducks", "swimming", "water",  "green", and the later stages select some keywords to form a semantically and grammatically correct caption gradually. 
	
	The difference between autoregressive decoding and masked non-autoregressive decoding during inference is that masked non-autoregressive decoding is naturally close to language generation of human beings. More specifically, human beings firstly generate keywords of the visual scene in brain, then choose other words to connect different pieces and compose the entire sentences obeying linguistic rules. It is a first visual then linguistic generation process, and the visual information lays the foundation of captions and linguistic information assists to form the final caption in a compositional manner rather than a sequential manner, which will be better at preserving meaningful semantic information. Moreover, $K$ stages of iterative selection and generation enable the inference process to obtain the deliberation ability to repeatedly adjust the caption according to the visual feature and generate a complete caption in the last stage. For example, in Fig.~\ref{fig1}, the first output caption contains "two ducks and ducks", then the second caption modifies to "two couple of ducks" and finally adjusts to "a couple of ducks" in the last captions, since only two ducks exist in the image. Last but not least, masked non-autoregressive decoding generates the entire sentence at one step, and thus the quality of the preceding tokens will not significantly influence that of the later ones, which fundamentally eases the sequential error accumulation existing in autoregressive decoding. However, masked non-autoregressive decoding also suffers from error accumulation between stages. By contrast, autoregressive decoding is a left-to-right and word-by-word generation process, and thus the generated tokens of the later step heavily depends on that of the preceding steps, which is prone to sequential error accumulation once the preceding tokens are inappropriate. Worse still, it has only one chance to generate the entire caption without the capability to adjust the preceding inappropriate tokens. Therefore, autoregressive decoding is reasonably good at maintaining fluency but difficult at accurately telling the rich salient semantic content of the images.
	
	\subsection{Inference rules}
	It is crucial to preserve the most informative tokens and mask other positions in the captions generated by each stage to generate a newly masked input sequence. In this paper, we adopt a straightforward approach. In this approach, those tokens that are not included in the token set of high frequency, as well as the tokens that are of high probability and not repetitive with the selected tokens thus far, are assigned as high preference to preserve. In Fig.~\ref{fig1}, for example, we preserve "two", "ducks", "swimming" and "water" of the first stage output sequence. Moreover, the generated captions of the last stage are processed by choosing tokens that are not repetitive with the previously selected ones. 
	
	Regarding the determination of the sequence lengths of captions during inference, we firstly calculate the distribution of different lengths of the training data, and then choose a random length $T$ for a caption from this distribution. Subsequently, the continual $T$ [MASK] tokens are fed to the model, such that a complete caption can be finally decoded. The other alternative is that we directly set a fixed sequence length for all images. The model will be automatically forced to generate coarse or fine captions depending on the length but with similar semantic information.
	
	Moreover, when the output captions of the last stage are masked with second largest ratio and fed into the model again to start a new round ($K-1$ stages), the generated captions could be better and finer in principle. The reason is that the model could start a new round with more accurate and informative masked sequences. Experimentally we observe that two rounds is slightly better than just one round, and the performance of more than two rounds tends to become saturated.
	\section{Experiments}
	\subsection{Dataset and implementation details}
	To evaluate our captioning model, we use the standard MSCOCO dataset \citep{lin2014microsoft}. We adopt the Karpathy split \citep{Karpathy2015Deep}, where $113,287$ images with $5$ captions for each image are used for training, $5000$ images are for validation and $5000$ images are for offline testing. We preprocess all captions, including lowercasing all captions, truncating captions longer than 16 words, and replacing words occurring no more than $5$ times with [UNK] token.
	In terms of implementation details, we use the visual features extracted from Faster R-CNN \citep{Ren2017Faster} in TopDown \citep{anderson2017bottom}. The decoder has $6$ layers and $8$ attention heads. The hidden size is set to $512$, and the feed-forward size is $2048$. Since the decoder stacks contain several layers, we calculate cross entropy losses of $2$nd, $4$th and $6$th layers and obtain the final average total loss with different weights, namely $1.0$, $1.1$, $1.2$. In this manner, we force the whole decoding stacks to converge to predict the target tokens, which speeds up the convergence. Adam optimizer \citep{kingma2014adam} is adopted along with warmup \citep{Vaswani2017Attention} to adjust the learning rate. During inference, the token set of high frequency is \{"a", "on"\}, which accounts for around $20\%$ of the whole tokens. 
	
	\subsection{Experimental settings}
	
	The model architectures of AIC, NAIC and MNIC are identical, as they all adopt the transformer decoder. In addition, in terms of different variants, we also conduct an extensive ablation study. More specifically, when  introducing random words into input sequences, the "Noise" tag is ticked. Moreover, we explore the performance of different combinations of masking percentages during training and inference respectively. All models use the same hyper-parameters and training strategy, and are evaluated using greedy decoding. 
	To compare with state-of-the-art captioning models, we also report the results of the models that are trained using cross entropy loss in their paper. Adaptive \citep{lu2017knowing}, TopDown~\citep{anderson2017bottom} and Skeleton \citep{Wang2017Skeleton} adopt LSTM as the decoder and use beam search, which is always better than greedy decoding. Adaptive and TopDown use different kinds of attention mechanisms. Skeleton decomposes the original caption into skeleton sentence and attributes using two LSTMs. CompCap \citep{Dai2018A} directly introduces a novel compositional paradigm to factorize semantics and syntax. 
	
	\begin{table}
		\caption{Performance comparisons with different evaluation metrics in offline testing. The masking ratio set of MNIC are all $\{ 0.4, 0.6, 0.8, 1.0 \}$ during training and inference, where $1$R and $2$R indicate first and second round during inference, respectively. The sequence length for all images of MNIC and NAIC is fixed at $11$, which occupies a large portion of length distribution. Latency is computed as the time to decode a single sentence without minibatching, and the values are averaged over the whole offline test set. The decoding is implemented on a single NVIDIA GeForce GTX 1080 Ti.}
		\label{table1}
		\centering
		\begin{tabular}{lcccccccccc}
			\toprule
			Models&  B1& B2& B3& B4& MT& RG& CD& SP& Latency&  SpeedUp\\
			\midrule
			Adaptive~\citep{lu2017knowing}&           74.2& \textbf{58.0}& \textbf{43.9}& 33.2& 26.6& -& 108.5& 19.5&  -& -\\
			TopDown~\citep{anderson2017bottom}&           \textbf{77.2}& -&    -&    \textbf{36.2}& 27.0& \textbf{56.4}& \textbf{113.5}& 20.3& -& -\\
			Skeleton~\citep{Wang2017Skeleton}&         74.2& 57.7& 44.0& 33.6& 26.8& 55.2& 107.3& 19.6& -& -\\
			CompCap~\citep{Dai2018A} &          -& -& -& 25.1& 24.3& 47.8& 86.2& 19.9& -& -\\
			\midrule
			AIC&           74.0& 57.3& 42.9& 31.8& 26.9& 54.7& 106.6& 20.2& 171ms& 1.00$\times$ \\
			NAIC&  72.5& 51.4& 34.2& 22.2& 22.6& 53.0& 79.7& 16.7& \textbf{18ms}& \textbf{9.50$\times$}\\
			\midrule
			MNIC ($1$R)& 75.4& 57.7& 42.6& 30.9& \textbf{27.5}& 55.6& 108.1& 21.0& 61ms& 2.80$\times$\\
			MNIC ($2$R)& 75.5& 57.9& 43.0& 31.5& \textbf{27.5}& 55.7& 108.5& \textbf{21.1}& 103ms&1.66$\times$\\
			\bottomrule
		\end{tabular}
	\end{table}
	
	\subsection{Experiment results}
	\paragraph{General comparisons}
	We compare the performance of different captioning models in the test portion of Karpathy splits of MSCOCO on BLEU 1,2,3,4 (B1, B2, B3, B4) ~\citep{papineni2002bleu}, METEOR (MT) \citep{banerjee2005meteor}, ROUGE (RG)~\citep{lin2004rouge}, CIDEr (CD)~\citep{vedantam2015cider} and SPICE (SP)~\citep{anderson2016spice}. In particular, SPICE focuses on semantic analysis and has higher correlation with human judgment, and other metrics favor frequent training n-grams and measure the overall sentence fluency, which are more preferred in sequential decoding based methods. As shown in Table~\ref{table1}, MNIC with different rounds obtain the best results on SP and MT, but is inferior to LSTM based autoregressive decoders in terms of some other metrics. AIC and TopDown use the same visual features and both adopt autoregressive decoding. However, TopDown is better than AIC in terms of all metrics, which demonstrates that the LSTM based decoder may be more suitable than the transformer based decoder for image captioning. However, though MNIC adopts the transformer based decoder, the SPICE score of MNIC largely surpasses that of TopDown and AIC, which indicates the compositional manner of MNIC can preserve semantic content more effectively than the sequential manner of TopDown and AIC. In terms of different decoding methods, masked non-autoregressive decoding of MNIC totally outperforms non-autoregressive decoding of NAIC and mainly outperforms autoregressive decoding of AIC except for a slightly drop on B4, which demonstrates that MNIC can address multimodality problem of non-autoregressive decoding and exhibits the advantage of the bidirectional language model over the unidirectional language model of AIC. By comparing different compositional paradigms, MNIC largely outperforms CompCap on all metrics, which indicates that MNIC can offer better consideration of both semantics and fluency of generated captions. 
	
	When comparing the latency regarding the models, it is shown that NAIC achieves a speedup of around a factor of 10 over AIC. More specifically, MNIC  ($1$R) and MNIC ($2$R) achieve factors of $2.80$ and $1.66$, respectively. It is worth mentioning that NAIC processes only using $1$ stage, and MNIC uses $4$ stages and $7$ stages respectively for $1$ round and $2$ round for ratio set $\{ 0.4, 0.6, 0.8, 1.0 \}$. As such, the smaller stage number $K$ is, the more significant speedup can be obtained.
	
	\begin{table}
		\caption{Ablation performance in offline testing. Different masking ratio sets are adopted during training and inference. The "Noise" tag is ticked if random words are introduced into input sequences during training. The sequence lengths of MNICs are randomly set using an identical random seed.}
		\label{table2}
		\centering
		\begin{tabular}{cllccccccc}
			\toprule
			No.& Training set&    Inference Set& Noise& B1& B4& MT& RG& CD& SP \\
			\midrule
			$1$& $\{ .4, .6, .8, 1. \}$&  $\{ .4, .6, .8, 1. \}$& & 73.0& 29.0& 26.8& 54.0& 102.4& 20.9\\
			\midrule
			$2$& $\{ .1, .3, .5, .7, 1. \}$&  $\{ .1, .3, .5, .7, 1. \}$&   $\surd$&   72.8& 28.7& 26.8& 53.8& 100.8& 20.9\\
			$3$& $\{ .2, .4, .6, .8, 1. \}$&  $\{ .2, .4, .6, .8, 1. \}$&  $\surd$& 73.0& 28.8& 26.9& 54.0& 102.0& 20.9\\
			$4$& $\{ .3, .5, .7, 1. \}$&  $\{ .3, .5, .7, 1. \}$&   $\surd$& 73.1& 29.2& 27.0& 54.4& 102.2& \textbf{21.0}\\
			$5$& $\{ .4, .6, .8, 1. \}$&  $\{ .4, .6, .8, 1. \}$&  $\surd$& 74.0& 30.2& \textbf{27.2}& \textbf{54.8}& \textbf{103.8}& \textbf{21.0}\\
			$6$& $\{ .5, .7, 1. \}$&  $\{ .5, .7, 1. \}$&  $\surd$&            74.0& 29.6& 26.8& 54.6& 101.7& 20.5\\
			$7$& $\{ .6, .8, 1. \}$&  $\{ .6, .8, 1. \}$&   $\surd$&           \textbf{75.2}& \textbf{30.5}& 26.4& \textbf{54.8}& 101.1& 20.2\\
			$8$& $\{ .7, 1. \}$&  $\{ .7, 1. \}$&  $\surd$&            74.5& 28.9& 25.6& 54.1& 97.3& 19.2\\
			\midrule
			$9$& $\{ .1, .3, .5, .7, 1. \}$& $\{ .4, .6, .8, 1. \}$&  $\surd$&       73.5& 29.7& 26.9& 54.4& 103.1& \textbf{21.0}\\
			$10$& $\{.7, 1. \}$&  $\{ .4, .6, .8, 1. \}$&  $\surd$& 73.7& 29.9& 26.7& 54.6& 101.2& 20.1\\
			\bottomrule
		\end{tabular}
	\end{table}
	\begin{figure}[t]
		\begin{tabular}{cccc}
			\subfigure[]{
				\begin{minipage}[]{1.1in}
					\centering
					\includegraphics[scale=0.132]{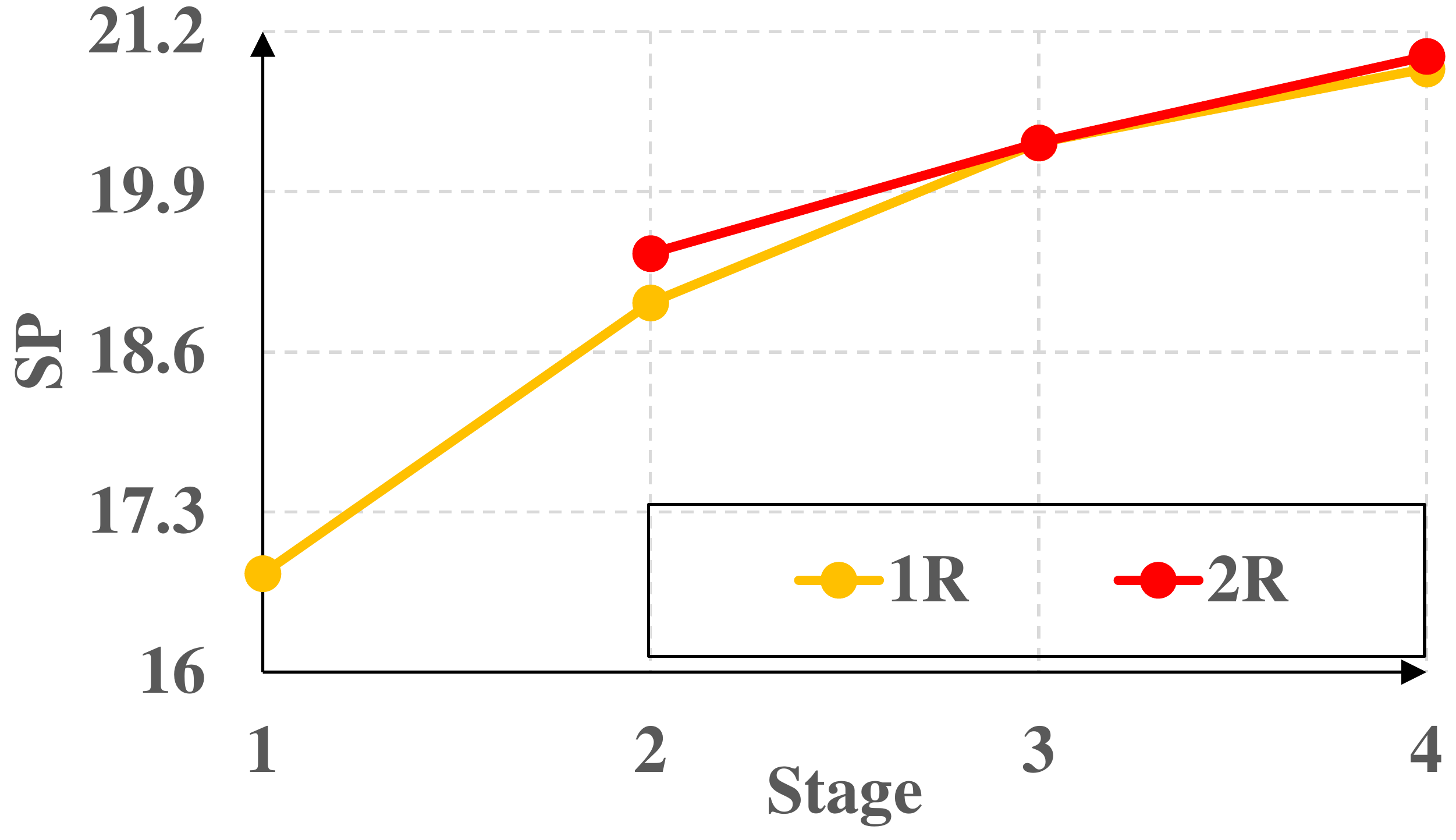}
				\end{minipage}
				\label{fig2:a}
			}&
			\subfigure[]{
				\begin{minipage}[]{1.1in}
					\centering
					\includegraphics[scale=0.27]{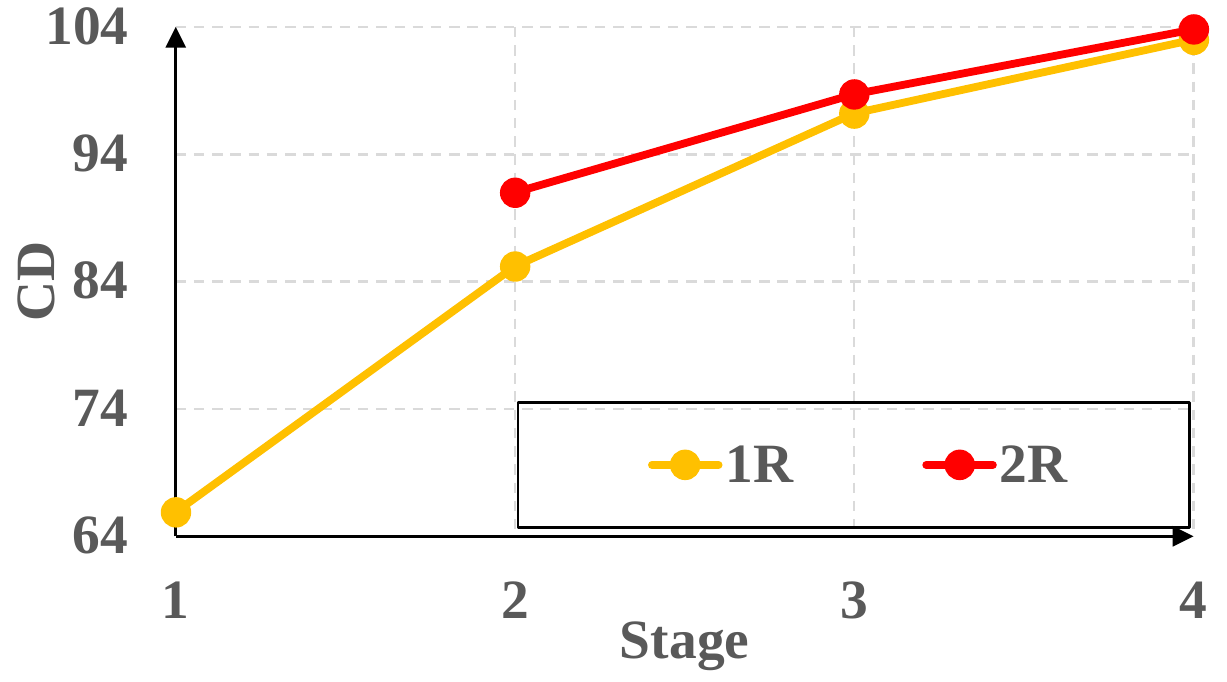}
				\end{minipage}
				\label{fig2:b}
			}&
			\subfigure[]{
				\begin{minipage}[]{1.1in}
					\centering
					\includegraphics[scale=0.20]{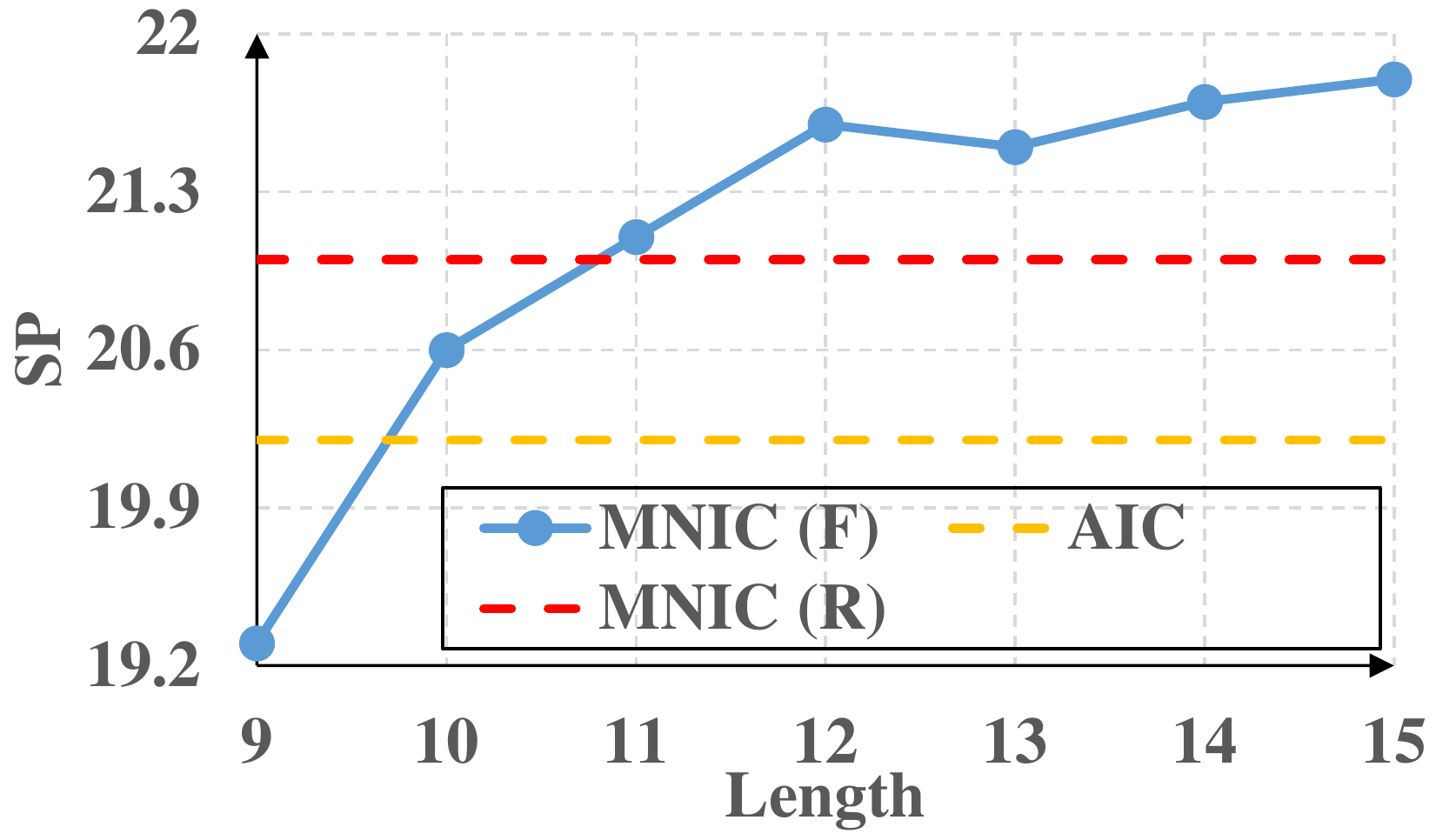}
				\end{minipage}
				\label{fig2:c}
			}&
			\subfigure[]{
				\begin{minipage}[]{1.1in}
					\centering
					\includegraphics[scale=0.20]{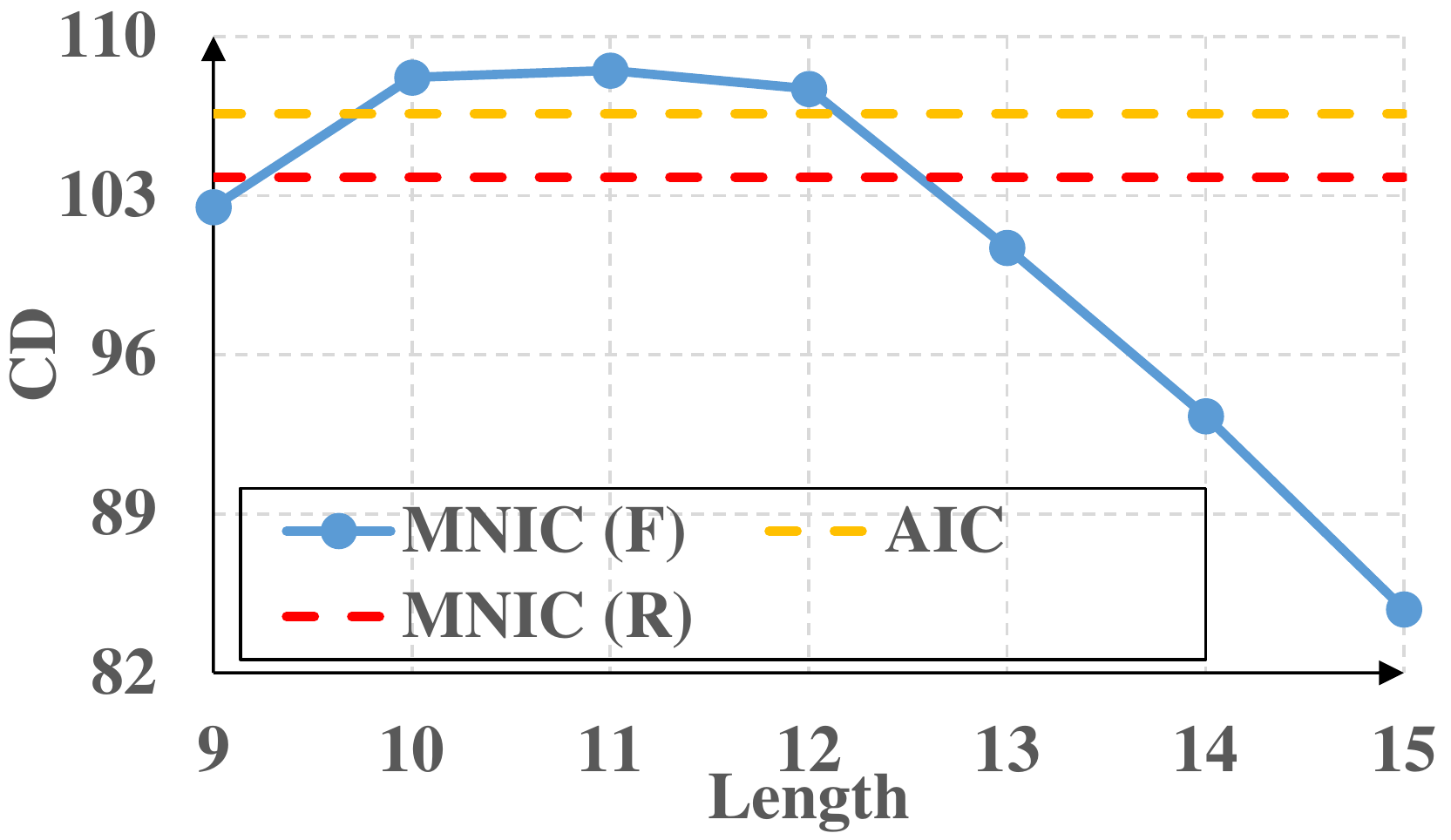}
				\end{minipage}
				\label{fig2:d}
			}
			\centering
		\end{tabular}
		\caption{Investigations of the influences of different stages and lengths in terms of SP and CD. 
		}
		\label{fig2}
	\end{figure}
	
	\paragraph{Ablation study and analysis}
	
	In Table~\ref{table2}, when comparing the experimental results of No. $1$ and No. $5$ , MNIC with Noise outperforms MNIC without Noise in terms of all metrics, since introducing random words into the input sequence can enhance the inference process and lead to significant performance improvement. 
	Regarding different combinations of masking ratios in Table~\ref{table2}, the configuration $\{ 0.4, 0.6, 0.8, 1.0 \}$ achieves the best comprehensive score.
	To investigate the influence of different smallest masking ratios, we conduct experiments as shown in No. $2$-$8$. It is interesting to find that the metric scores first increase and reach a comprehensive peak at No. $5$, after which the performance decreases. The results suggest that when the smallest masking ratio in the ratio set is intermediate, the ratio set achieves the best performance, and two extremes perform even worse. The results could be influenced by the ratio set towards training, as well as the ratio set towards inference. Hence, we additionally conduct experiment as shown in No. $9$-$10$. With the same inference ratio set of No. $5$ and No. $9$, the training set $\{ 0.4, 0.6, 0.8, 1.0 \}$ outperforms $\{ 0.1, 0.3, 0.5, 0.7, 1.0 \}$, which indicates the former learns a better bidirectional language model than the latter. With the same training set of No. $2$ and No. $9$, the inference set $\{ 0.4, 0.6, 0.8, 1.0 \}$ outperforms $\{ 0.1, 0.3, 0.5, 0.7, 1.0 \}$, which indicates the former helps to generate sequences of higher quality than the latter. It is the same case if the smallest ratio in the ratio set is sufficiently large as in No. $10$ .
	
	\begin{table}[t]
		\caption{Illustration of the diversity of different methods from different aspects, where the scores of Skeleton are extracted from \citep{Wang2017Skeleton}, and TopDown \citep{anderson2017bottom} and CompCap \citep{Dai2018A} are extracted from \citep{Dai2018A}. The masking ratio set of MNIC is $\{ 0.4, 0.6, 0.8, 1.0 \}$ and the sequence lengths are randomly set.}
		\label{table3}
		\centering
		\begin{tabular}{lccccc}
			\toprule
			&          Skeleton~\citep{Wang2017Skeleton}& TopDown~\citep{anderson2017bottom}& CompCap~\citep{Dai2018A}& AIC& MNIC\\
			\midrule
			Novel Caption ($\%$)& 52.24& 45.05& \textbf{90.48}& 71.16&  82.54\\
			Unique Caption ($\%$)& 66.96& 61.58& 83.86& 79.32&  \textbf{91.62}\\
			Vocabulary Usage ($\%$)& -&             7.97& 9.18&   \textbf{12.53}&  11.62\\
			\bottomrule
		\end{tabular}
	\end{table}
	
	\begin{figure}
		\begin{center}
			\begin{tabular}{ll}
				\toprule
				\multirow{4}{*}{\includegraphics[height=0.6in, width=1.1in]{COCO_val2014_000000272212.jpg}}&
				GT: Cows grazing on the grass in front of a building.\\&
				AIC: three cows grazing in a field.\\&
				MNIC1: the cows are grazing in the green field.\\&
				MNIC2: a group of cows standing on top of a lush green hillside.
				\\
				\midrule
				\multirow{4}{*}{\includegraphics[height=0.6in, width=1.1in]{COCO_val2014_000000198641.jpg}}&
				GT: A black cat laying on a white lap top.\\&
				AIC: a cat laying on top of a computer desk.\\&
				MNIC1: a cat is sitting on a computer desk.\\&
				MNIC2: a black cat laying on a desk in front of a computer monitor.
				\\
				\midrule 
				\multirow{4}{*}{\includegraphics[height=0.6in, width=1.1in]{COCO_val2014_000000524064.jpg}}&
				GT: A couple of men riding horses down a street with tall buildings.\\&
				AIC: a couple of horses down a street.\\&
				MNIC1: a couple of men riding horses down a city street.\\&
				MNIC2: a couple of men riding on the backs of horses down a street.
				\\
				\bottomrule
			\end{tabular}
			\caption{Example of ground truth captions, the generated captions of AIC and MNIC using different sequence lengths.}
			\label{fig3}
		\end{center}
	\end{figure}
	\paragraph{Performance in different stages}
	In Fig.~\ref{fig2:a} and Fig.~\ref{fig2:b}, we report the results of each stage in $2$ rounds of MNIC with the setting No.~$5$ in Table~\ref{table2}, where the output sequences of the last stage of round 1 (1R) is masked with ratio $0.8$ and then fed to the model to start round 2 (2R). The scores within round 1 or round 2 continually improve across different stages in terms of both metrics, which indicates that the generated sequences of early stages are poor in quality but can be continually adjusted and improved in the later stages due to the bidirectional language model. It is also worth noting that the $2$nd stage of round 1 and round 2 has the same input masking ratio, but the latter largely outperforms the former. This implies that regarding the generated sequences of the last stage of round 1, more important semantic information is injected into the model than those of the first stage of round 1, even though most tokens are masked. The final result of round 2 is only slightly superior to round 1, suggesting that only one round can also achieve a good trade-off between performance and computational complexity, which coincides with the observations in Table~\ref{table1}. 
	
	\paragraph{Performance in different sequence lengths}
	In Fig.~\ref{fig2:c} and Fig.~\ref{fig2:d}, we compare the performances of different fixed sequence lengths (denoted as MNIC (F)) with random sequence lengths (denoted as MNIC (R)) and AIC, where the ratio set of MNIC is $\{ 0.4, 0.6, 0.8, 1.0 \}$. In general, MNIC (F) outperforms MNIC (R) and AIC in terms of SP and CD. Fig.~\ref{fig2:c} suggests that when the sequence length increases, SP score can be continually improved, which demonstrates that when the sequence length is short, MNIC tends to generate a coarse caption that contains the most salient semantic content, in analogous to AIC. On the other hand, MNIC will generate finer captions that contain more semantic information if long sequence length is set, as illustrated in quantitative examples in Fig.~\ref{fig3}.  On the contrary, CD score favors intermediate sequence length, since it is easier for intermediate length to generate the syntactically and semantically correct captions. 
	
	\paragraph{Diversity study}
	Since masked non-autoregressive decoding is a first visual then linguistic generation process and can better preserve semantics, it is contrary to autoregressive decoding that more diverse captions are expected to be generated. To analyze the diversity of generated captions, we compute novel caption percentage, unique caption percentage and vocabulary usage, which respectively account for the percentage of captions that have not been seen in the training data, the percentage of captions that are unique in the whole generated captions and the percentage of words in the vocabulary that are adopted to generate captions. 
	In Table~\ref{table3}, it is obvious that MNIC and CompCap can achieve better results, which suggest that compositional methods can generate more diverse captions than sequential methods, such as Skeleton \citep{Wang2017Skeleton}, TopDown \citep{anderson2017bottom} and AIC, though Skeleton decomposes into the skeleton and attribute LSTM to generate a caption. 
	\section{Conclusions}
	In this paper, we propose a novel decoding method for image captioning. In contrast to the typical methods which generate captions using autoregressive and non-autoregressive decoding, the novelty of this paper lies in that the masked language model is trained by masking certain ratios of the input data, and the captions are generated in a compositional instead of sequential manner. 
	As such, the proposed masked non-autoregressive decoding can effectively tackle the issues including sequential error accumulation, slow generation, improper semantics, lack of diversity and multimodality problem arising from the typical autoregressive and non-autoregressive decoding.  
	The experimental results provide evidence regarding the effectiveness and efficiency of the proposed scheme. 
	
	
	\small{
		\bibliographystyle{unsrtnat} 
		\bibliography{egbib}
	}

	\clearpage
	\begin{center}
		\large{\textbf{Supplementary Material}}
	\end{center}
	
\begin{figure}[h]
		\centering
		\begin{tabular}{c}
			\subfigure{
				\begin{minipage}{2.4in}
					\centering
					\includegraphics[scale=0.4]{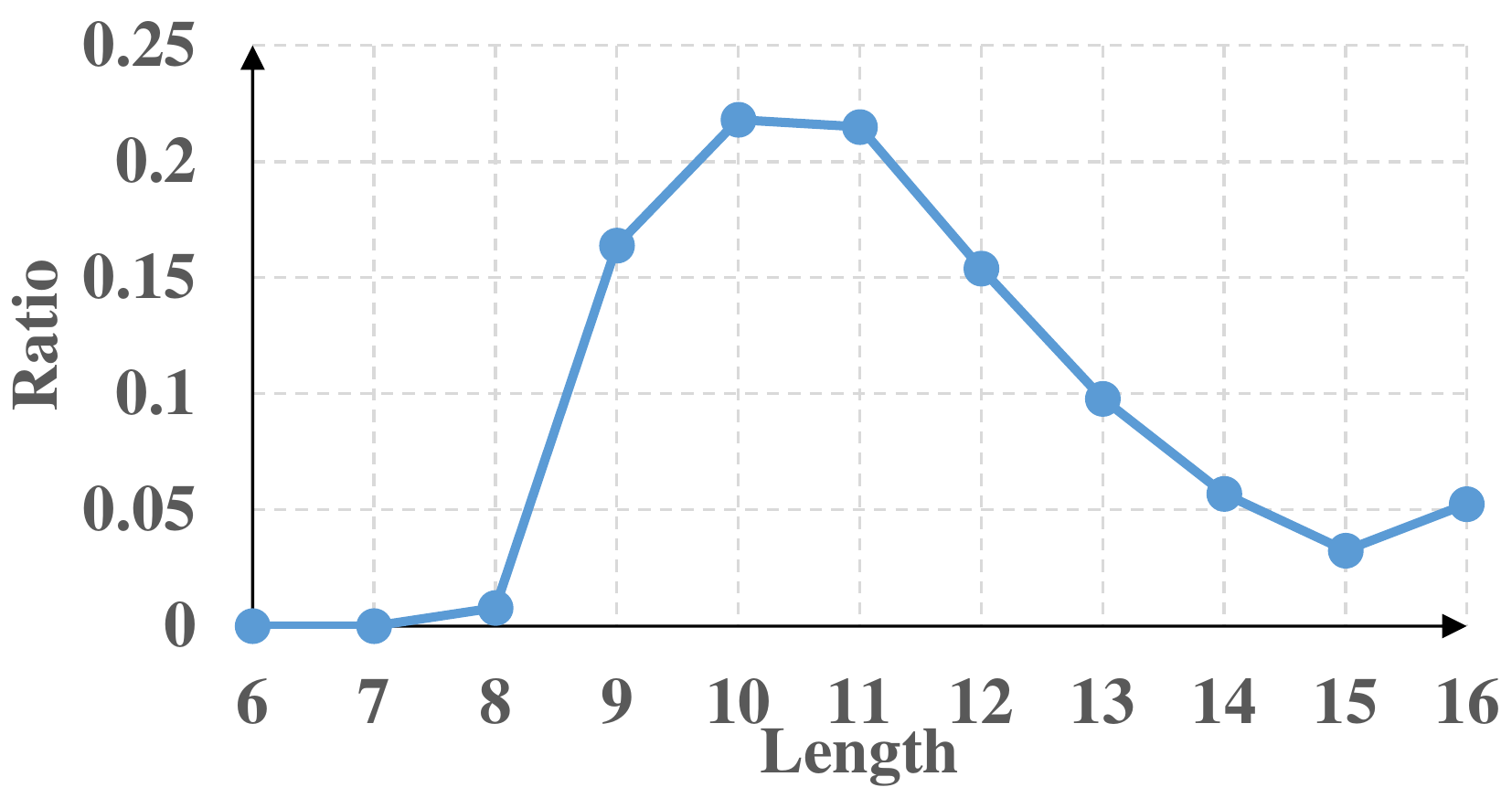}
				\end{minipage}
			}
		\end{tabular}
		\caption{The distribution of sequence length of the training captions. 
		}
		\label{fig4}
	\end{figure}

\begin{figure}[t]
		\begin{tabular}{cc}
			\subfigure[]{
				\begin{minipage}[]{2.4in}
					\centering
					\includegraphics[scale=0.4]{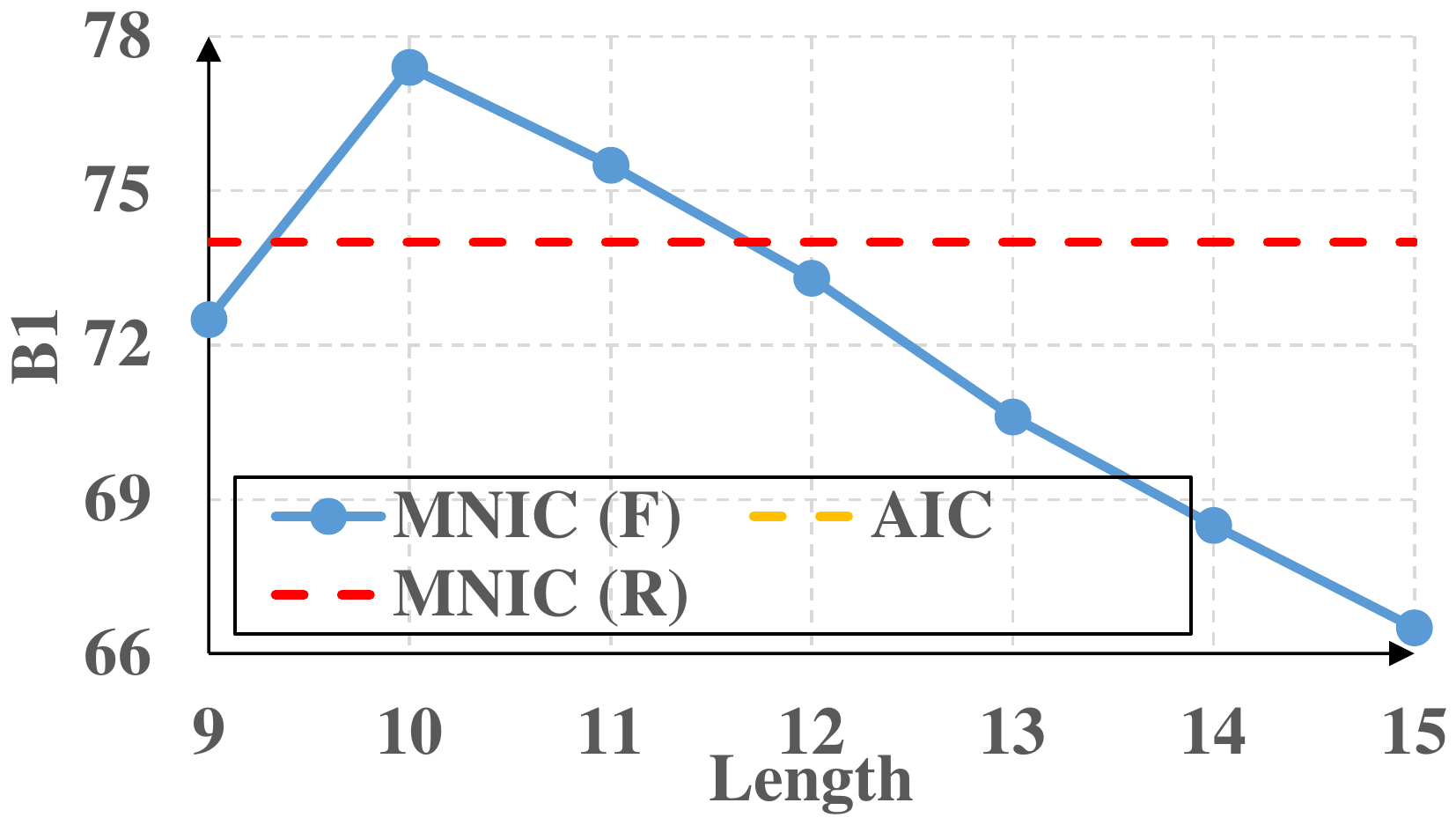}
				\end{minipage}
				\label{fig5:a}
			}&
			\subfigure[]{
				\begin{minipage}[]{2.4in}
					\centering
					\includegraphics[scale=0.4]{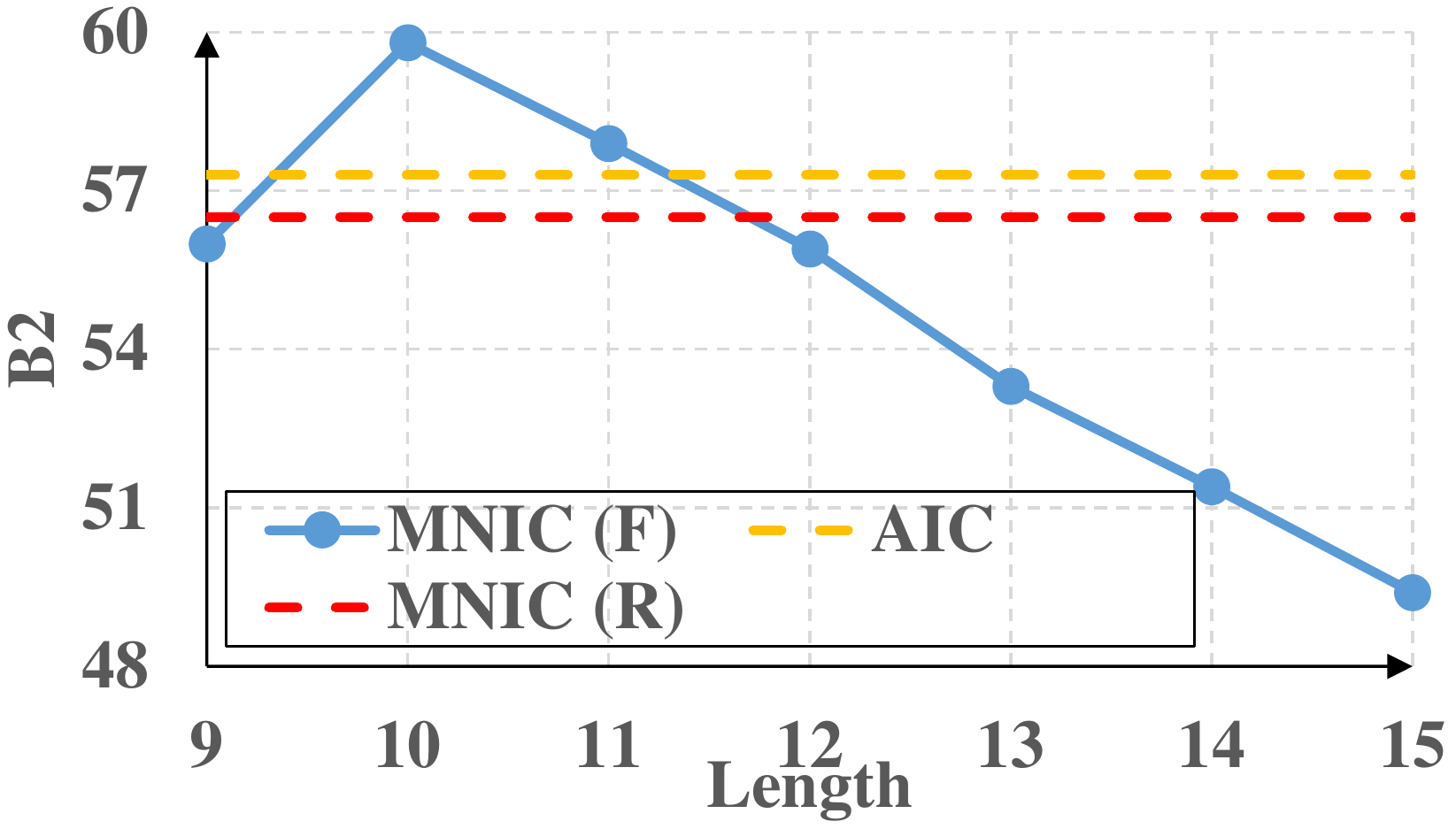}
				\end{minipage}
				\label{fig5:b}
			}\\
			\subfigure[]{
				\begin{minipage}[]{2.4in}
					\centering
					\includegraphics[scale=0.4]{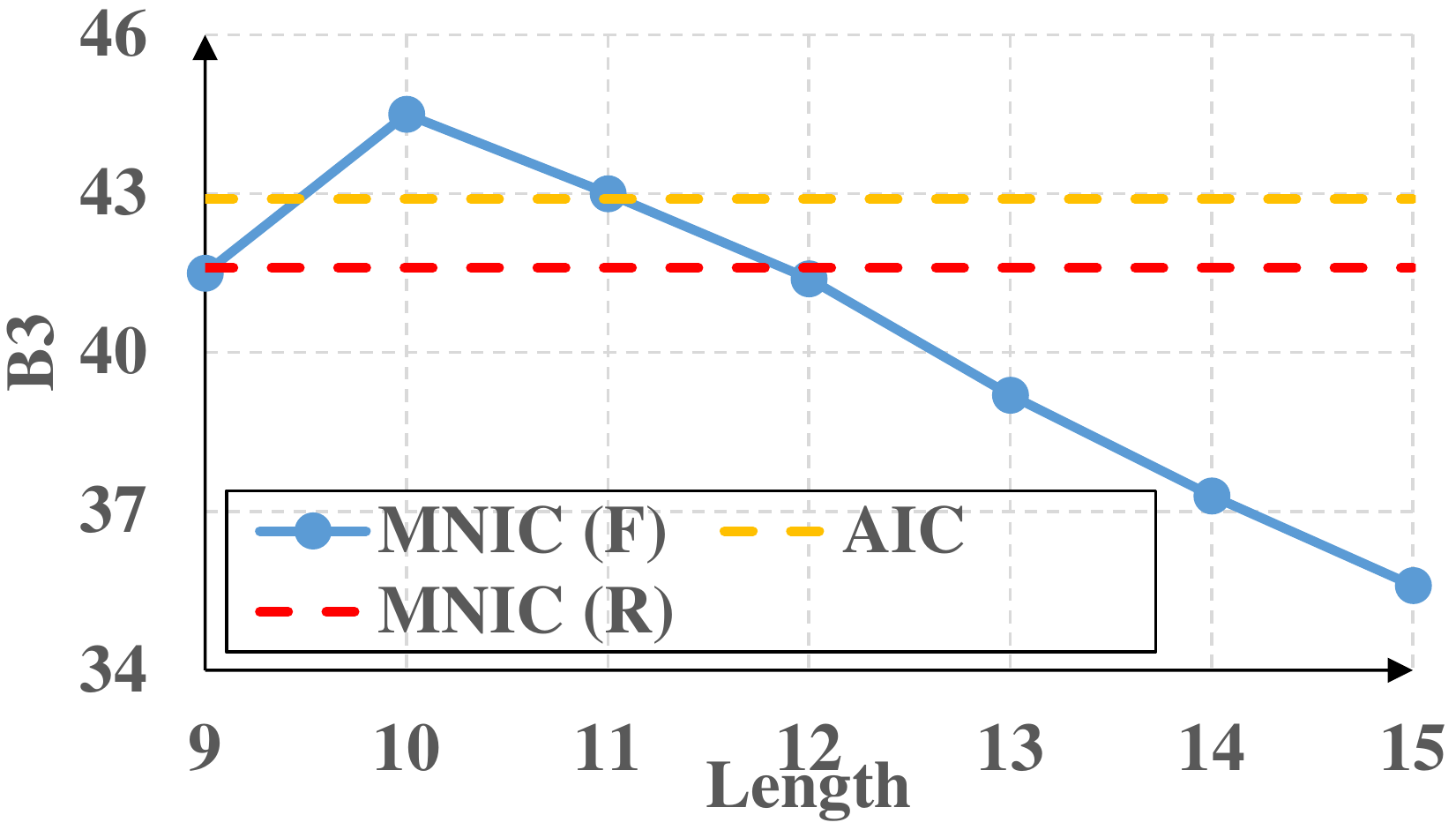}
				\end{minipage}
				\label{fig5:c}
			}&
			\subfigure[]{
				\begin{minipage}[]{2.4in}
					\centering
					\includegraphics[scale=0.4]{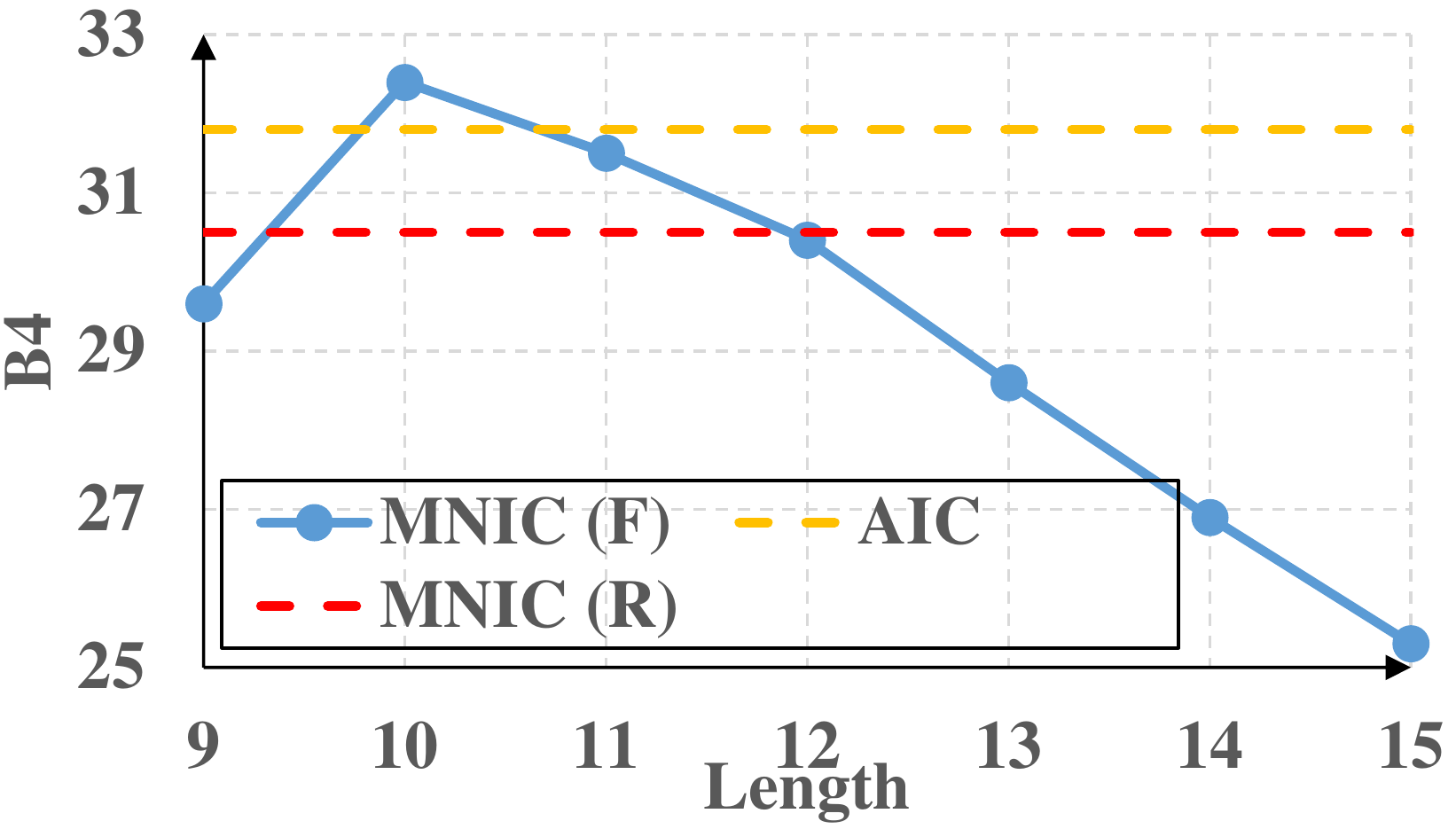}
				\end{minipage}
				\label{fig5:d}
			}\\
			\subfigure[]{
				\begin{minipage}[]{2.4in}
					\centering
					\includegraphics[scale=0.4]{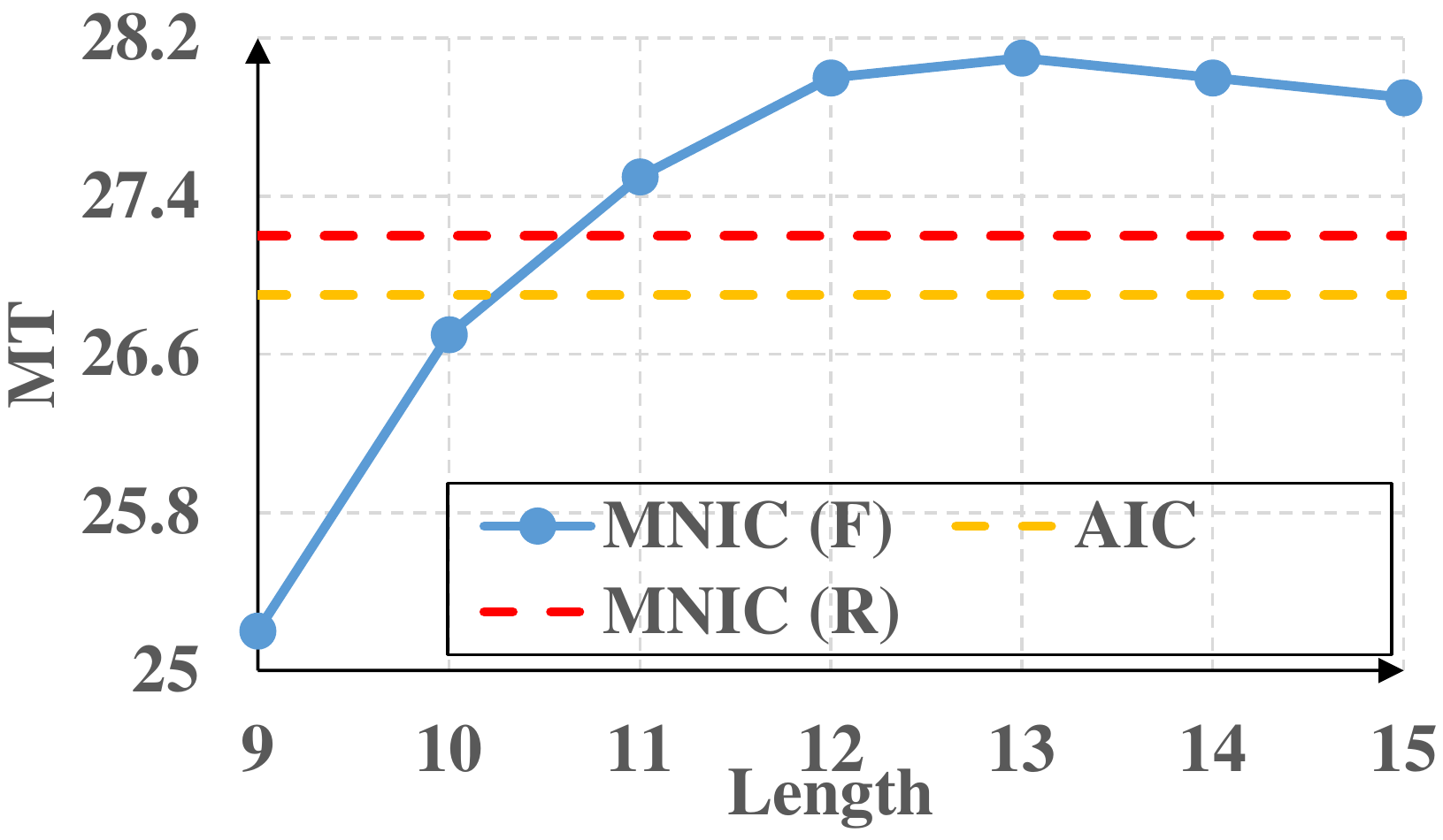}
				\end{minipage}
				\label{fig5:e}
			}&
			\subfigure[]{
				\begin{minipage}[]{2.4in}
					\centering
					\includegraphics[scale=0.4]{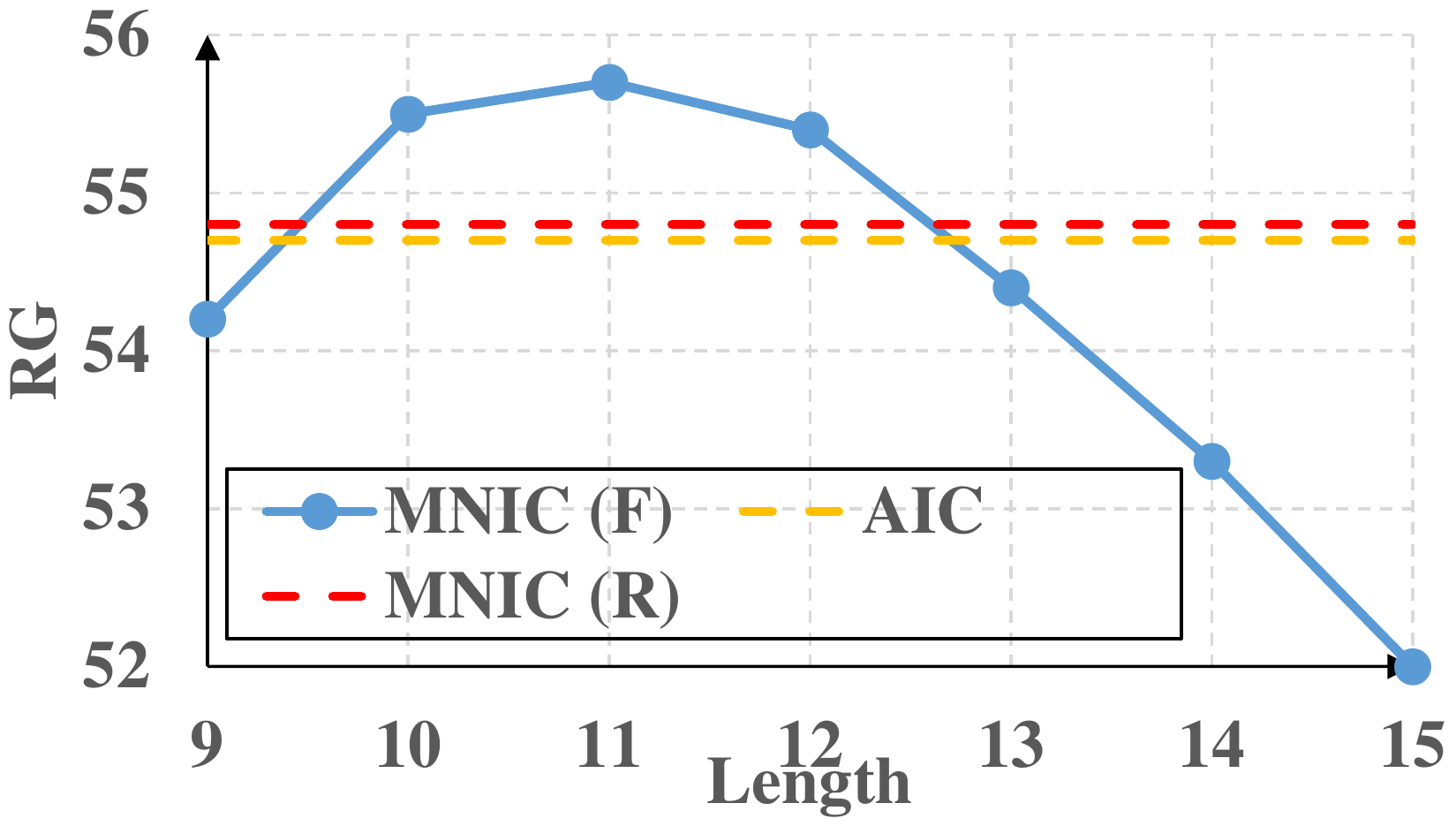}
				\end{minipage}
				\label{fig5:f}
			}\\
			\subfigure[]{
				\begin{minipage}[]{2.4in}
					\centering
					\includegraphics[scale=0.4]{LENGTH_CD.pdf}
				\end{minipage}
				\label{fig5:g}
			}&
			\subfigure[]{
				\begin{minipage}[]{2.4in}
					\centering
					\includegraphics[scale=0.4]{LENGTH_SP.pdf}
				\end{minipage}
				\label{fig5:h}
			}\\
			\centering
		\end{tabular}
		\caption{We extend scores of SP and CD of Fig.\ref{fig2:c} and Fig.\ref{fig2:d} in the manuscript to those of the whole evaluation metrics.}
		\label{fig5}
	\end{figure}

\begin{figure}[t]
		\begin{center}
			\begin{tabular}{ll}
				\toprule
				\multirow{9}{*}{\includegraphics[height=0.6in, width=1.1in]{COCO_val2014_000000272212.jpg}}&
				GT: Cows grazing on the grass in front of a building.\\&
				AIC: three cows grazing in a field.\\&
				~9~: the cows are grazing in the green field.\\&
				10: a group of cows are in a green field.\\&
				11: a herd of cattle standing on a lush green hillside.\\&
				12: a herd of cattle standing on top of a green hillside.\\&
				13: a group of cows standing on top of a lush green hillside.\\&
				14: three cows grazing in a grassy field with a house in the background.\\&
				15: three cows grazing in a green field with a white house in the background.
				\\
				\midrule
				\multirow{9}{*}{\includegraphics[height=0.6in, width=1.1in]{COCO_val2014_000000281455.jpg}}&
				GT: A group of birds flying over the water.\\&
				AIC: a flock of birds flying over a body of water.\\&
				~9~: a flock of birds flying through the air.\\&
				10: a bunch of birds flying in the air.\\&
				11: a flock of birds flying through a cloudy sky.\\&
				12: a flock of birds flying through the air over a beach.\\&
				13: a flock of birds flying through the air on a cloudy day.\\&
				14: a flock of birds flying through the air on a partly cloudy day.\\&
				15: a flock of birds flying in the air with a clouds in the sky.
				\\
				\midrule 
				\multirow{9}{*}{\includegraphics[height=0.6in, width=1.1in]{COCO_val2014_000000128675.jpg}}&
				GT: A man holding one water ski at the lake.\\&
				AIC: a man is flying a kite in the water.\\&
				~9~: a person riding a surfboard in the water.\\&
				10: a man is riding a board in the water.\\&
				11: a man that is on a surfboard in the water.\\&
				12: a man riding a surfboard on a boat in the water.\\&
				13: a man riding a surf board on a boat in the water.\\&
				14: a man riding a surfboard in the water with trees in the background.\\&
				15: a man riding a surfboard in the water with a trees in the background.
				\\
				\midrule 
				\multirow{9}{*}{\includegraphics[height=0.6in, width=1.1in]{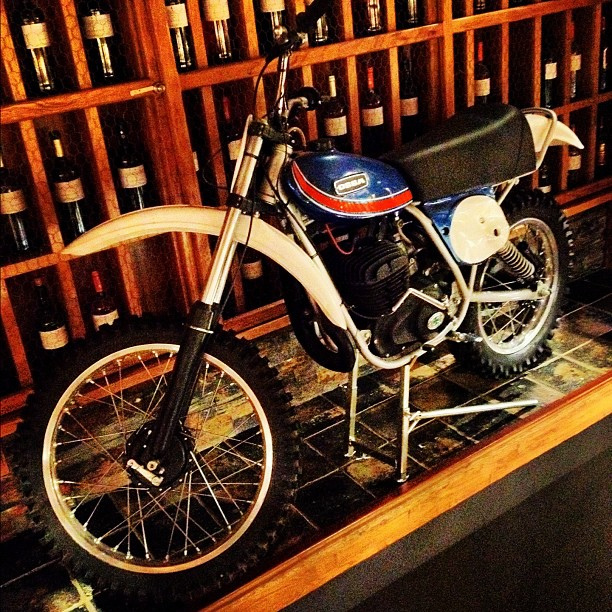}}&
				GT: A bike sits among bottles of wine on a shelf.\\&
				AIC: a motorcycle parked on a wooden floor next to a bookshelf.\\&
				~9~: a motorcycle is parked in a building.\\&
				10: a motorcycle parked on a shelf in a building.\\&
				11: a motorcycle parked on a wooden shelf in a building.\\&
				12: a motorcycle parked in front of a shelf in a shelves.\\&
				13: a motorcycle parked in front of a wooden shelf in a shelves.\\&
				14: a motorcycle parked on a wooden floor in front of some wooden shelves.\\&
				15: a motorcycle parked in front of a shelf next to a bunch of shelves.
				\\
				\midrule 
				\multirow{9}{*}{\includegraphics[height=0.6in, width=1.1in]{COCO_val2014_000000033759.jpg}}&
				GT: A boy swinging a baseball bat at a game.\\&
				AIC: a young boy swinging a bat towards a ball.\\&
				~9~: a young boy is swinging at a baseball.\\&
				10: a young boy swinging a bat at a ball.\\&
				11: a young boy swinging a baseball bat on a field.\\&
				12: a young boy holding a bat on top of a field.\\&
				13: a young boy holding a baseball bat prepares to hit a ball.\\&
				14: a young boy holding a baseball bat on top of a baseball field.\\&
				15: a young boy swinging a baseball bat at a ball on a baseball field.
				\\
				\bottomrule
			\end{tabular}
			\caption{Example of ground truth captions, the generated captions of AIC and MNIC using different sequence lengths (from $9$ to $15$). Since the generated captions of the last stage are processed by choosing tokens that are not repetitive with the previously selected one, the sequence lengths is not necessarily strictly equal to the predefined lengths.}
			\label{fig6}
		\end{center}
	\end{figure}

	\paragraph{Influence of different sequence lengths}
	Here, we further analyze the influence of different sequence lengths both during training and inference. We report the distribution of sequence length of training captions in Fig.\ref{fig4}, which demonstrates that the ratios of different lengths are uneven, and $10$ and $11$ account for a large part. It will affect the quality of language models for different lengths. As illustrated in Fig.\ref{fig5}, the figures of B1,2,3,4, RG, CD scores highly corresponds to Fig.\ref{fig4}. However, the uneven length distribution will not influence MT and SP scores in Fig.\ref{fig5:e} and Fig.\ref{fig5:h}, since captions of longer lengths can contain more and finer semantic information. Furthermore, more qualitative results of generated captions of different sequence lengths, which coincide with these observations and interpretations, are shown in Fig.\ref{fig6}.
	
\end{document}